# Algorithms and Complexity Results for Exact Bayesian Structure Learning

**Sebastian Ordyniak** and **Stefan Szeider**
Institute of Information Systems
Vienna University of Technology, Austria

## Abstract

Bayesian structure learning is the NP-hard problem of discovering a Bayesian network that optimally represents a given set of training data. In this paper we study the computational worst-case complexity of exact Bayesian structure learning under graph theoretic restrictions on the super-structure. The super-structure (a concept introduced by Perrier, Imoto, and Miyano, JMLR 2008) is an undirected graph that contains as subgraphs the skeletons of solution networks. Our results apply to several variants of score-based Bayesian structure learning where the score of a network decomposes into local scores of its nodes.

*Results:* We show that exact Bayesian structure learning can be carried out in non-uniform polynomial time if the super-structure has bounded treewidth and in linear time if in addition the super-structure has bounded maximum degree. We complement this with a number of hardness results. We show that both restrictions (treewidth and degree) are essential and cannot be dropped without loosing uniform polynomial time tractability (subject to a complexity-theoretic assumption). Furthermore, we show that the restrictions remain essential if we do not search for a globally optimal network but we aim to improve a given network by means of at most $k$ arc additions, arc deletions, or arc reversals ($k$-neighborhood local search).

*Keywords:* Bayesian structure learning, super-structure, treewidth, fixed-parameter tractability, parameterized complexity

## 1 Introduction

Bayesian structure learning is the important task of discovering a Bayesian network that represents a given set of training data. Unfortunately the problem is NP-hard (Chickering 1996). This predicament has motivated a wide range of approaches, including heuristics-based algorithms that compute near-optimal solutions (see, e.g., Heckerman, Geiger, & Chickering 1995; Chickering 2003). In recent years several exponential-time algorithms for *exact* Bayesian structure learning have been proposed (see, e.g., Parviainen & Koivisto 2009; Perrier, Imoto, & Miyano 2008; Silander & Myllymäki 2006). Recent progress has been made to limit the space requirement by advanced dynamic programming techniques (Parviainen & Koivisto 2009) and to limit the exponential time requirement by restricting the search to networks whose skeletons are subgraphs of a given undirected graph that specifies a *"super-structure"* (Perrier, Imoto, & Miyano 2008). Recent research indicates that the super-structure can be practically computed and effectually used to guide the search for near-optimal Bayesian networks (Mukund & Jeff 2004; Anton & Carlos 2007; Pieter, Daphne, & Andrew 2006).

In this paper we study the worst-case time complexity of exact Bayesian structure learning under graph-theoretic restrictions on the super-structure. In particular, we consider bounds on the *treewidth* and on the *maximum degree* of super-structures.

Our results are as follows:

(1) Exact Bayesian structure learning is feasible in *non-uniform polynomial* time if the treewidth of the super-structure is bounded by an arbitrary constant.

(2) Exact Bayesian structure learning is feasible in *linear time* if both treewidth and maximum degree of the super-structure are bounded by arbitrary constants.

By "non-uniform" we mean that the order of the polynomial depends on the treewidth. We obtain results (1) and (2) by means of a dynamic programming algorithm along a decomposition tree of the super-structure.

We show that—in a certain sense—both results are optimal:

(3) Exact Bayesian structure learning for instances with super-structures of maximum degree 4 (but unbounded treewidth) is not feasible in polynomial time unless P = NP. Thus, in (1) and (2) we cannot drop the bound on the treewidth.

Research supported by the European Research Council, grant reference 239962.

(4) Exact Bayesian structure learning for instances with super-structures of bounded treewidth (but unbounded maximum degree) is not feasible in uniform polynomial time unless FPT = W[1]. Thus, in (2) we cannot drop the bound on the degree.

FPT $\neq$ W[1] is a widely accepted complexity theoretic assumption (Downey & Fellows 1999). For example, FPT = W[1] implies the (unlikely) existence of a $2^{o(n)}$ algorithm for $n$-variable 3SAT (Impagliazzo, Paturi, & Zane 2001; Flum & Grohe 2006). Result (3) easily follows from Chickering's reduction (Chickering 1996). We establish result (4) by means of a parameterized reduction from a variant of the Maximum Clique problem. We will provide necessary background on parameterized complexity and parameterized reductions in Section 2.2.

We further extend the hardness results (3) and (4) from the search for an optimal network to the presumably easier problem of improving a given network by changing at most $k$ of its arcs (with the operations of arc addition, arc deletion, and arc reversal). We refer to this restricted problem as $k$-*neighborhood local search* or $k$-*local search* for short. By trivial reasons $k$-local search is feasible in non-uniform polynomial time $n^{O(k)}$. We show, however, that uniform polynomial-time tractability is again unlikely:

(5) $k$-local search for instances with super-structures of bounded maximum degree is not possible in uniform polynomial time unless FPT = W[1].

(6) $k$-local search for instances with super-structures of bounded treewidth is not possible in uniform polynomial time unless FPT = W[1].

We obtain result (5) by a reduction from the Red/Blue Non-Blocker problem (Downey & Fellows 1999). If both the maximum degree and the treewidth are bounded, then $k$-local search is feasible in linear time, however this result is subsumed by (2). Both hardness results (5) and (6) even hold for several cases where not all of the three operations (addition, deletion, reversal) are available, for example if arc reversal is the only operation.

## 2  Preliminaries

In this section we will introduce the basic concepts and notions that we will use throughout the paper.

### 2.1  Basic Graph Theory

We will assume that the reader is familiar with basic graph theory. We consider undirected graphs and directed graphs (digraphs). A *DAG* is a directed acyclic graph. We write $V(G) = V$ and $E(G) = E$ for the sets of vertices and edges of a (directed or undirected) graph $G = (V, E)$. We denote an undirected edge between vertices $u$ and $v$ as $\{u, v\}$ and a directed edge (or arc), directed from $u$ to $v$ as $(u, v)$. For a subset $V' \subseteq V$ we write $G[V']$ to denote the *induced subgraph* $G' = (V', E')$ where $E' = \{\, e \subseteq V' : e \in E \,\}$ if $G$ is undirected and $E' = \{\, e \in V' \times V' : e \in E \,\}$ if $G$ is directed. If $G$ is a digraph we define $P_G(v) = \{\, u \in V(G) : (u, v) \in E(G) \,\}$ as the set of *parents* of $v$ in $G$. An undirected graph $G' = (V, E')$ is the *skeleton* of $G$ if $E' = \{\, \{u, v\} : (u, v) \in E(G) \,\}$.

### 2.2  Parameterized Complexity

Parameterized complexity provides a theoretical framework to distinguish between uniform and non-uniform polynomial-time tractability with respect to a parameter. An instance of a parameterized problem is a pair $(I, k)$ where $I$ is the *main part* and $k$ is the *parameter*; the latter is usually a non-negative integer. A parameterized problem is *fixed-parameter tractable* if there exist a computable function $f$ and a constant $c$ such that instances $(I, k)$ of size $n$ can be solved in time $O(f(k)n^c)$. FPT is the class of all fixed-parameter tractable decision problems. Fixed-parameter tractable problems are also called *uniform polynomial-time tractable* because if $k$ is considered constant, then instances with parameter $k$ can be solved in polynomial time where the order of the polynomial is independent of $k$ (in contrast to non-uniform polynomial-time running times such as $n^k$).

Parameterized complexity offers a completeness theory similar to the theory of NP-completeness. One uses *parameterized reductions* which are many-one reductions where the parameter for one problem maps into the parameter for the other. More specifically, problem $L$ reduces to problem $L'$ if there is a mapping $R$ from instances of $L$ to instances of $L'$ such that (i) $(I, k)$ is a yes-instance of $L$ if and only if $(I', k') = R(I, k)$ is a yes-instance of $L'$, (ii) $k' = g(k)$ for a computable function $g$, and (iii) $R$ can be computed in time $O(f(k)n^c)$ where $f$ is a computable function, $c$ is a constant, and $n$ denotes the size of $(I, k)$. The parameterized complexity class W[1] is considered as the parameterized analog to NP. For example, the parameterized Maximum Clique problem (given a graph $G$ and a parameter $k \geq 0$, does $G$ contain a complete subgraph on $k$ vertices?) is W[1]-complete under parameterized reductions. Note that there exists a trivial non-uniform polynomial-time $n^k$ algorithm for the Maximum Clique problems that checks all sets of $k$ vertices.

### 2.3  Tree Decompositions

Treewidth is an important graph parameter that indicates in a certain sense the "tree-likeness" of a graph.

The treewidth of a graph $G = (V, E)$ is defined via the following notion of decomposition: a *tree decomposition* of $G$ is a pair $(T, \chi)$ where $T$ is a tree and $\chi$ is a labeling function with $\chi(t) \subseteq V$ for every tree node $t$, such that the following conditions hold:

1. Every vertex of $G$ occurs in $\chi(t)$ for some tree node $t$.
2. For every edge $\{u, v\}$ of $G$ there is a tree node $t$ such that $u, v \in \chi(t)$.
3. For every vertex $v$ of $G$, the tree nodes $t$ with $v \in \chi(t)$ induce a connected subtree of $T$.

The *width* of a tree decomposition $(T, \chi)$ is the size of a largest set $\chi(t)$ minus 1 among all nodes $t$ of $T$. A tree decomposition of smallest width is *optimal*. The *treewidth* of a graph $G$, denoted $\mathrm{tw}(G)$, is the width of an optimal tree decomposition of $G$.

Given $G$ with $n$ vertices and a constant $w$, it is possible to decide whether $G$ has treewidth at most $w$, and if so, to compute an optimal tree decomposition of $G$ in time $O(n)$ (Bodlaender 1996). Furthermore there exist powerful heuristics to compute tree decomposition of small width in a practically feasible way (Gogate & Dechter 2004).

## 3 Bayesian Structure Learning

In this section we define the theoretical framework for Bayesian structure learning that we shall use for our considerations. We closely follow the abstract framework used by Parviainen and Koivisto (2009) which encloses a wide range of score-based approaches to structure learning. We assume that the input data specifies a set $V$ of *nodes* (or variables) and a *local score function* $f$ that assigns to each $v \in V$ and each subset $A \subseteq V \setminus \{v\}$ a non-negative real number $f(v, A)$. Given the local score function $f$, the problem is to find a DAG $D = (V, E)$ such that the *score* of $D$ under $f$

$$f(D) := \sum_{v \in V} f(v, \mathrm{P}_D(v))$$

is as large as possible (the DAG $D$ together with certain local probability distributions forms a Bayesian network). This setting accommodates several popular scores like BDe, BIC and AIC (Parviainen & Koivisto 2009; Chickering 1995).

We consider the following decision problem:

EXACT BAYESIAN STRUCTURE LEARNING
*Instance:* A local score function $f$ defined on a set $V$ of nodes, a real number $s > 0$.
*Question:* Is there a DAG $D$ such that $f(D) \geq s$?

For our complexity theoretic considerations we will assume that the local score function $f$ is given as the list of all tuples $(v, A, f(v, A))$ for $v \in V$ and $A \subseteq V \setminus \{v\}$ where $f(v, A) > 0$. We define $\mathcal{P}_f(v) := \{P \subseteq V : f(v, P) > 0\} \cup \{\emptyset\}$ to be the set of all *potential parent sets* of $v$. We also define

$$\delta_f := \max_{v \in V} |\mathcal{P}_f(v)|;$$

which will be an important measurement for our worst-case analysis of running times.

Let $f$ be a local score function defined on a set $V$ of nodes. The *super-structure* of $f$ is the undirected graph $S_f = (V, E_f)$ where $E_f$ contains an edge $\{u, v\}$ if and only if $u$ is a potential parent of $v$, i.e., if $u \in P$ for some $P \in \mathcal{P}_f(v)$.

We say that a DAG $D$ is *admissible* for $f$ if the skeleton of $D$ is a spanning subgraph of the super-structure $S_f$. Furthermore, we say that a DAG $D$ is *strictly admissible* for $f$ if for every vertex $v \in V(D)$ we have $\mathrm{P}_D(v) \in \mathcal{P}_f(v)$. Note that every strictly admissible DAG is also admissible. Furthermore, there always exists a (strictly) admissible DAG $D$ with the highest score: If $D$ is not (strictly) admissible, i.e., if there exists $v \in V(D)$ such that $f(v, \mathrm{P}_D(v)) = 0$, we can delete all arcs $(w, v)$ such that $w \in \mathrm{P}_D(v)$. This does not decrease the score since $f(v, \emptyset) \geq f(v, \mathrm{P}_D(v)) = 0$ for every such $v$.

## 4 An Algorithm for Exact Bayesian Structure Learning

In this section we present the dynamic programming algorithm and establish our tractability results. For the remainder of this section $w$ denotes an arbitrary but fixed constant.

**Theorem 1.** *Given a local score function $f$ with a super-structure $S_f = (V, E_f)$ of treewidth bounded by a constant $w$. Then we can find in time $O(\delta_f^{w+1} \cdot |V|)$ a DAG $D$ with maximal score $f(D)$.*

**Corollary 1.** EXACT BAYESIAN STRUCTURE LEARNING *can be decided in polynomial time for instances where the super-structure has bounded treewidth. The problem can be decided in linear time if additionally the super-structure has bounded maximum degree.*

*Proof.* The first statement follows immediately from the theorem since $\delta_f$ is bounded by the total input size of the instance. The second statement follows since $\delta_f$ is bounded whenever the maximum degree $d$ of the super-structure is bounded as clearly $\delta_f \leq 2^d$. □

We are going to establish Theorem 1 by means of a dynamic programming algorithm along a tree decomposition for $S_f$, computing local information at the nodes of the tree decomposition that can then be put together to form an optimal DAG. For this approach, it is convenient to consider tree decompositions in the following normal form (Kloks 1994): A triple $(T, \chi, r)$ is a *nice tree decomposition* of a graph $G$ if $(T, \chi)$ is a tree decomposition of $G$, the tree $T$ is rooted at node $r$, and each node of $T$ is of one of the following four types:

1. a *leaf node*: a node having no children;

2. a *join node*: a node $t$ having exactly two children $t_1, t_2$, and $\chi(t) = \chi(t_1) = \chi(t_2)$;
3. an *introduce node*: a node $t$ having exactly one child $t'$, and $\chi(t) = \chi(t') \cup \{v\}$ for a vertex $v$ of $G$;
4. a *forget node*: a node $t$ having exactly one child $t'$, and $\chi(t) = \chi(t') \setminus \{v\}$ for a vertex $v$ of $G$.

For convenience we will also assume that $\chi(r) = \emptyset$ for the root $r$ of $T$. For a nice tree decomposition $(T, \chi, r)$ we define $\chi^*(t)$ to be the union of all the sets $\chi(t')$ where $t'$ is contained in the subtree of $T$ rooted at $t$.

Given a tree decomposition of a graph $G$ of width $w$, one can effectively obtain in time $O(|V(G)|)$ a nice tree decomposition of $G$ with $O(|V(G)|)$ nodes and of width at most $w$ (Kloks 1994).

In the following we will assume that we are given an instance $I = (V, f)$ of EXACT BAYESIAN STRUCTURE LEARNING together with a nice tree decomposition $(T, \chi, r)$ for $S_f$ of width at most $w$.

A *partial solution* for a tree node $t \in V(T)$ is a digraph that can be obtained as the induced subdigraph $D[\chi^*(t)]$ of a strictly admissible DAG $D$ for $f$. For a tree node $t$ let $\mathcal{D}(t)$ denote the set of all partial solutions for $t$. For a partial solution $D \in \mathcal{D}(t)$ we set

$$f_t(D) = \sum_{v \in (V(D) \setminus \chi(t))} f(v, \mathrm{P}_D(v)),$$

i.e., $f_t(D)$ is the sum of the scores of all nodes of $D$ except for the nodes in $\chi(t)$.

A *record* of a tree node $t \in V(T)$ is a triple $R = (a, p, s)$ such that:

1. $a$ is a mapping $\chi(t) \to \mathcal{P}_f(v)$;
2. $p$ is a transitive binary relation on $\chi(t)$;
3. $s$ is a non-negative real number.

We say that a record *represents* a partial solution $D \in \mathcal{D}(t)$ if it satisfies the following conditions:

1. $a(v) \cap V(D) = \mathrm{P}_D(v)$ for every $v \in \chi(t)$.
2. For every pair of vertices $v_1, v_2 \in \chi(t)$ it holds that $(v_1, v_2) \in p$ if and only if $D$ contains a directed path from $v_1$ to $v_2$.

We say that a record $R = (a, p, s)$ of a tree node $t \in V(T)$ is *valid* if it represents some DAG $D \in \mathcal{D}(t)$ and $s$ is the maximum score $f_t(D)$ over all DAGs in $\mathcal{D}(t)$ represented by $R$. With each tree node $t \in V(T)$ we associate the set $\mathcal{R}(t)$ of all valid records representing partial solutions in $\mathcal{D}(t)$.

In a certain sense, $\mathcal{R}(t)$ is a succinct representation of the optimal elements of $\mathcal{D}(t)$, using space that only depends on $w$ and $\delta_f$, but not on $|V|$.

The next three lemmas will allow us to compute the valid records of a tree node from the valid records of its children.

**Lemma 1** (join nodes). *Let $t_1$, $t_2$ be the children of $t$ in $T$. Then $\mathcal{R}(t)$ can be computed from $\mathcal{R}(t_1)$ and $\mathcal{R}(t_2)$ in time $O(\delta_f^{w+1})$.*

*Proof.* It follows from the above definitions that a record $R = (a, p, s)$ of $t$ is valid if and only if there are valid records $R_1 = (a_1, p_1, s_1) \in \mathcal{R}(t_1)$ and $R_2 = (a_2, p_2, s_2) \in \mathcal{R}(t_2)$ such that:

1. $a = a_1 = a_2$.
2. $p$ is the transitive closure of $p_1 \cup p_2$.
3. $p$ is irreflexive, i.e., there is no $v \in \chi(t)$ such that $(v, v) \in p$.
4. $s = s_1 + s_2$.

It follows that $\mathcal{R}(t)$ can be computed by considering all pairs of records $R_1 \in \mathcal{R}(t_1)$ and $R_2 \in \mathcal{R}(t_2)$ and checking conditions 1–4. Since, there are at most $O(\delta_f^{w+1})$ valid records for every $t \in V(T)$ and for every such pair the time required to check the conditions only depend on $w$, the result follows. □

**Lemma 2** (introduce node). *Let $t$ be an introduce node with child $t'$, such that $\chi(t) = \chi(t') \cup \{v_0\}$. Then $\mathcal{R}(t)$ can be computed from $\mathcal{R}(t')$ in time $O(\delta_f^{w+1})$.*

*Proof.* A record $R = (a, p, s)$ of $t$ is valid if and only if there is a set $P \in \mathcal{P}_f(v_0)$ and a valid record $R' = (a', p', s') \in \mathcal{R}(t')$ such that:

1. $a(v_0) = P$.
2. For every $v \in \chi(t')$ it holds that $a(v) = a'(v)$.
3. $p$ is the transitive closure of the relation $p' \cup \{(u, v_0) : u \in P\} \cup \{(v_0, u) : v_0 \in a'(u), u \in \chi(t')\}$.
4. $p$ is irreflexive.
5. $s = s'$.

It follows that $\mathcal{R}(t)$ can be computed by checking for every pair $(P, R')$ as defined above, whether it satisfies conditions 1–5. Since there are at most $\delta_f$ possible sets $P$ and at most $O(\delta_f^w)$ possible valid records for $t'$ (observe that $|\chi(t')| \leq w$) the result follows from the fact that for every pair $(P, R')$ the conditions can be checked in time that only depends on $w$. □

**Lemma 3** (forget node). *Let $t$ be a forget node with child $t'$ such that $\chi(t) = \chi(t') \setminus \{v_0\}$. Then $\mathcal{R}(t)$ can be computed from $\mathcal{R}(t')$ in time $O(\delta_f^{w+1})$.*

*Proof.* A record $R = (a, p, s)$ of $t$ is valid if and only if there is a valid record $R' = (a', p', s') \in \mathcal{R}(t')$ such that:

1. $a$ and $p$ are the restrictions of $a'$ and $p'$ to $\chi(t)$, respectively. That is, $a(u) = a'(u)$ for all $u \in \chi(t)$, and $p = \{(u, v) \in p' : u, v \in \chi(t)\}$.
2. $s = s' + f(v_0, a'(v_0))$.

Evidently $\mathcal{R}(t)$ can be computed from $\mathcal{R}(t')$ in time $O(\delta_f^{w+1})$. □

We are now ready to prove Theorem 1.

*Proof.* Let $I = (V, f)$ be an instance of EXACT BAYESIAN STRUCTURE LEARNING where the super-structure $S_f$ has treewidth $w$ (a constant) and $|V| = n$. We compute a nice tree decomposition $(T, \chi, r)$ of $S_f$ of width $w$ and with $O(n)$ nodes. This can be accomplished in time $O(n)$ (see the discussion in Section 2.3).

Next we compute the sets $\mathcal{R}(t)$ via a bottom-up traversal of $T$. For a leaf node $t$ we can compute $\mathcal{R}(t)$ just by considering all valid records for every possible strictly admissible DAG on the at most $w + 1$ vertices in $\chi(t)$. We can now use Lemmas 1, 2 and 3 to compute the sets $\mathcal{R}(t)$ for all other $O(n)$ tree nodes in time $O(\delta_f^{w+1} \cdot n)$.

Since $\chi(r) = \emptyset$, the partial solutions for the root $r$ of $T$ are exactly the strictly admissible DAGs for $f$, and we have $f_r(D) = f(D)$ for each such DAG $D$. After the computation of the sets $\mathcal{R}(t)$ for all tree nodes $t$, the set $\mathcal{R}(r)$ contains exactly one record $R = (\emptyset, \emptyset, s)$. By the above considerations, it follows that $s$ is the largest score of all strictly admissible DAGs for $f$, and, as noted in Section 3, this is also the largest score of any DAG whose vertices belong to $V$. It is now easy to compute a DAG $D$ with score $f(D) = s$ via a top-down traversal of $T$ starting from $r$ and using the information previously stored at each node in $T$. This can also be accomplished in time $O(\delta_f^{w+1} \cdot n)$. □

## 5 Hardness Results

**Theorem 2** (Chickering 1996). EXACT BAYESIAN STRUCTURE LEARNING *is* NP-*hard for instances with super-structures of maximum degree* 4.

*Proof.* This theorem follows from Chickering's proof, we only sketch the argument. The reduction is from FEEDBACK ARC SET (FAS). The problem asks whether a digraph $D = (V, E)$ can be made acyclic by deleting at most $k$ arcs (the deleted arcs form a feedback arc set of $D$). The problem is NP-hard for digraphs with skeletons of maximum degree 4 (Karp 1972). Given an instance $(D, k)$ of FAS, where the skeleton of $D$ has maximum degree 4, we construct a set $V' = V(D) \cup E(D)$ of nodes and a local score function $f$ on $V'$ by setting $f((u,v), \{u\}) = 1$ for all $(u, v) \in E(D)$, $f(v, \{ (u,v) : u \in \mathrm{P}_D(v) \}) = |\mathrm{P}_D(v)|$ for all $v \in V(D)$, and $f(v, P) = 0$ in all other cases. Clearly, the super-structure $S_f$ is the undirected graph obtained from the skeleton of $D$ after subdividing every edge once, hence the maximum degree of $S_f$ is at most 4. It is easy to see that $D$ has a feedback arc set of size $\leq k$ if and only if there exists a DAG $D'$ whose skeleton is a spanning subgraph of $S_f$ with $f(D') \geq 2 \cdot |E| - k$. □

**Theorem 3.** EXACT BAYESIAN STRUCTURE LEARNING *parameterized by the treewidth of the super-structure is* W[1]-*hard.*

*Proof.* We devise a parameterized reduction from the following problem, which is well-known to be W[1]-complete (Pietrzak 2003).

PARTITIONED CLIQUE

*Instance:* A $k$-partite graph $G = (V, E)$ with partition $V_1, \ldots, V_k$ such that $|V_i| = |V_j| = n$ for $1 \leq i < j \leq k$.

*Parameter:* The integer $k$.

*Question:* Are there vertices $v_1, \ldots, v_k$ such that $v_i \in V_i$ for $1 \leq i \leq k$ and $\{v_i, v_j\} \in E$ for $1 \leq i < j \leq k$? (The graph $K = (\{v_1, \ldots, v_k\}, \{ \{v_i, v_j\} : 1 \leq i < j \leq k \})$ is a $k$-*clique* of $G$.)

Let $G = (V, E)$ be an instance of this problem with partition $V_1, \ldots, V_k$, $|V_1| = \cdots = |V_k| = n$. Let $\alpha = k^2 - 1$ and $\epsilon = 2k$. We construct a set $N$ of nodes and a local score function $f$ on $N$ such that (i) $\mathrm{tw}(S_f) \leq k(k-1)/2$ and (ii) $G$ has a $k$-clique if and only if there exists a DAG $D$ such that $f(D) \geq k(n-1)\alpha + (k(k-1)/2)\epsilon$. See Figure 1 for an illustration.

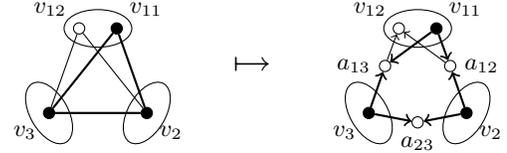

Figure 1: Illustration for the reduction in the proof of Theorem 3, $k = 3$.

We set $A = \{ a_{ij} : 1 \leq i < j \leq k \}$, $N = V(G) \cup A$, and $A_i = \{ a_{lk} : l = i \text{ or } k = i \}$ for every $1 \leq i \leq k$. We are now ready to define $f$. We set $f(v, A_i) = \alpha$ for every $v \in V_i$, and $f(a_{ij}, \{u, w\}) = \epsilon$ for every $1 \leq i < j \leq k$, $u \in V_i$, $w \in V_j$, and $\{u, w\} \in E(G)$. Furthermore we set $f(v, P) = 0$ for all the remaining combinations of $v$ and $P$. It is easy to see claim (i) as deleting the $k(k-1)/2$ vertices $a_{ij}$ from $S_f$ yields a collection of isolated vertices, i.e., a graph of treewidth 0. Hence, it remains to show claim (ii).

So suppose that $G$ has a $k$-clique $K = (\{v_1, \ldots, v_k\}, E_K)$, such that $v_i \in V_i$ for every $1 \leq i \leq k$. It follows that for the DAG $D$ with arc set $E(D) = \{ (v_i, a) : 1 \leq i \leq k,\ a \in A_i \} \cup \{ (a, v) : 1 \leq i \leq k,\ a \in A_i,\ v \in V_i \setminus \{v_i\} \}$ the following holds:

1. $f(v, \mathrm{P}_D(v)) = 0$, for every $v \in V(K)$;
2. $f(v, \mathrm{P}_D(v)) = \alpha$, for every $v \in V(G) \setminus V(K)$;
3. $f(a, \mathrm{P}_D(a)) = \epsilon$, for every $a \in A$.

Hence, $f(D) = k(n-1)\alpha + (k(k-1)/2)\epsilon$ and the only-if direction of claim (ii) follows.

To show the if direction of claim (ii) suppose that there exists a DAG $D$ such that $f(D) \geq k(n-1)\alpha + (k(k-1)/2)\epsilon$. It can be shown that such a score can only be obtained if every vertex in $A$ attains its maximum score and exactly one vertex $v_i$ from every $V_i$ does not. It is then easy to see that the vertices $\{v_1, \ldots, v_k\}$ form a $k$-clique in $G$ and the claim follows. □

Note that in contrast to Theorem 2, it is essential for Theorem 3 that the super-structure has unbounded degree: if both degree and treewidth are bounded then the problem is fixed-parameter tractable by Corollary 1 and so unlikely to be W[1]-hard.

## 6 $k$-Neighborhood Local Search

Important and widely used algorithms for Bayesian structure learning are based on local search methods (Heckerman, Geiger, & Chickering 1995). Usually the local search algorithm tries to improve the score of a given DAG by transforming it into a new DAG by adding, deleting, or reversing an arc (in symbols ADD, DEL, and REV, respectively). The main obstacle for local search methods is the danger of getting stuck at a poor local optimum. A possibility for decreasing this danger is to perform $k > 1$ elementary changes in one step, known as $k$-*neighborhood local search* or $k$-*local search* for short. For Bayesian structure learning, when we try to improve the score of a DAG on $n$ nodes, the $k$-local search space is of order $n^{O(k)}$. Therefore, if carried out by brute-forth, $k$-local search is too costly even for small values of $k$. It is therefore not surprising that most practical local search algorithms for Bayesian structure learning consider 1-neighborhoods only.

In this section we investigate whether under restrictions on the super-structure where EXACT BAYESIAN STRUCTURE LEARNING remains hard (as considered in Theorems 2 and 3) at least $k$-local search becomes easier compared to the general unrestricted case. Our results are mostly negative. In fact, somewhat surprisingly, $k$-local search remains hard even if edge reversal is the only allowed operation.

Before we give the hardness proofs we define $k$-local search more formally. Let $k \geq 0$ and $\mathbb{O} \subseteq \{\text{ADD}, \text{DEL}, \text{REV}\}$. Consider a DAG $D = (V, E)$. A directed graph $D' = (V', E')$ is a $k$-$\mathbb{O}$-*neighbor* of $D$ if

1. $D'$ is a DAG,
2. $V = V'$,
3. $E'$ can be obtained from $E$ by performing at most $k$ operations from the set $\mathbb{O}$.

For $\mathbb{O} \subseteq \{\text{ADD}, \text{DEL}, \text{REV}\}$ we consider the following parameterized decision problem.

$k$-$\mathbb{O}$-LOCAL SEARCH BAYESIAN STRUCTURE LEARNING

*Instance:* A local score function $f$, a DAG $D$ that is admissible for $f$, and an integer $k$.

*Question:* Is there a $k$-$\mathbb{O}$-neighbor $D'$ of $D$ with a higher score than $D$?

Note that the problem does not change if we require $D'$ to be admissible, as we can always avoid the addition of an inadmissible arc.

**Theorem 4.** *If $\mathbb{O} = \{\text{ADD}\}$ or $\mathbb{O} = \{\text{DEL}\}$, then $k$-$\mathbb{O}$-LOCAL SEARCH BAYESIAN STRUCTURE LEARNING is solvable in polynomial time.*

*Proof.* We only consider $\mathbb{O} = \{\text{ADD}\}$ as the proof for $\mathbb{O} = \{\text{DEL}\}$ is analogous. Let $I = (D, f, k)$ be the given instance of $k$-$\{\text{ADD}\}$-LOCAL SEARCH BAYESIAN STRUCTURE LEARNING. Note that there exists a $k$-$\{\text{ADD}\}$-neighbor $D'$ of $D$ with $f(D') > f(D)$ if and only if there exists a vertex $v \in V(D)$ such that the addition of at most $k$ incoming arcs increases the score of $v$ and the resulting digraph remains acyclic. Now, for every entry $f(v, P)$ such that $P \subseteq V(D) \setminus \{v\}$ one can easily check whether $f(v, P) > f(v, \mathrm{P}_D(v))$ and whether $P$ can be obtained from $\mathrm{P}_D(v)$ via the addition of at most $k$ incoming arcs such that the resulting digraph is acyclic. □

In view of Theorem 4 let us define a set $\mathbb{O} \subseteq \{\text{ADD}, \text{DEL}, \text{REV}\}$ to be *non-trivial* if $\mathbb{O} \notin \{\emptyset, \{\text{ADD}\}, \{\text{DEL}\}\}$.

**Theorem 5.** *Let $\mathbb{O} \subseteq \{\text{ADD}, \text{DEL}, \text{REV}\}$ be non-trivial. Then $k$-$\mathbb{O}$-LOCAL SEARCH BAYESIAN STRUCTURE LEARNING is W[1]-hard for parameter $\mathrm{tw}(S_f) + k$.*

*Proof.* We slightly modify the reduction given in the proof of Theorem 3. Let $D$ be the directed acyclic graph with vertex set $N$, arc set $\{(a, v) : a \in A_i, v \in V_i\}$. We set $k' = k(k-1)/2$. Then, for every $\mathbb{O}$ that contains the operation REV, it is easy to see—using the same arguments as in the proof of Theorem 3—that $G$ has a $k$-clique if and only if $D$ has a $k'$-$\mathbb{O}$-neighbor $D'$ with $f(D') > f(D)$. Similarly, for the remaining case $\mathbb{O} = \{\text{ADD}, \text{DEL}\}$, one can show that $G$ has a $k$-clique if and only if $D$ has a $2k'$-$\mathbb{O}$-neighbor $D'$ with $f(D') > f(D)$. □

**Theorem 6.** *Let $\mathbb{O} \subseteq \{\text{ADD}, \text{DEL}, \text{REV}\}$ be non-trivial. Then $k$-$\mathbb{O}$-LOCAL SEARCH BAYESIAN STRUCTURE LEARNING is W[1]-hard for parameter $k$, hardness even holds if the super-structure $S_f$ has bounded maximum degree.*

*Proof.* We devise a parameterized reduction from the following problem which is known to be W[1]-complete for every constant $d \geq 3$ (Downey & Fellows 1999; Flum & Grohe 2006).

BOUNDED DEGREE RED/BLUE NON-BLOCKER

*Instance:* An undirected graph $G = (V, E)$ with maximum degree $d$, where $V$ is the disjoint union of sets Red and Blue, and an integer $k$.

*Parameter:* The integer $k$.

*Question:* Is there a set $S \subseteq \text{Red}$ of size $k$ such that every $v \in \text{Blue}$ has a neighbor outside of $S$? ($S$ is a $k$-*Red/Blue non-blocker* of $G$).

Let $(G, \text{Red}, \text{Blue}, k)$ be an instance of this problem with $\text{Red} = \{v_r^1, \ldots, v_r^n\}$ and $\text{Blue} = \{v_b^1, \ldots, v_b^m\}$. We may assume that $G$ is bipartite with partition $\{\text{Red}, \text{Blue}\}$. To see this, observe that without affecting the answer we can remove every edge in $G$ between two vertices in Red and similarly we can remove every vertex in Blue that has a neighbor in Blue.

Let $k' = (d+1)k + 1$. We construct a DAG $D$ and a local score function $f$ such that $G$ has a $k$-Red/Blue non-blocker if and only if $D$ has a $k'$-$\mathbb{O}$-neighbor $D'$ with a higher score than $D$. The construction given below applies to all cases with REV $\in \mathbb{O}$. For the only remaining nontrivial set $\mathbb{O} = \{\text{DEL}, \text{ADD}\}$ it is easy to adapt the construction by setting $k'$ to $2((d+1)k+1)$.

To make the following arguments easier, it is convenient that all vertices in Red are of degree exactly $d$. Hence we introduce an intermediate graph $G'$ that is obtained from $G$ by adding $d - d(v)$ vertices for every $v \in \text{Red}$ and connecting each of these vertices by an edge to the corresponding $v$.

The DAG $D$ is obtained from $G'$ by applying the following steps (see Figure 2 for an illustration):

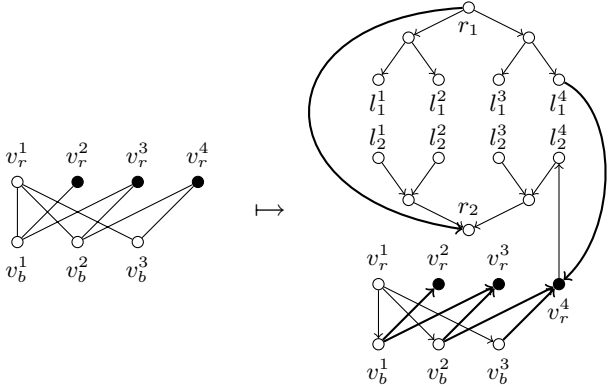

Figure 2: Illustration for the reduction in the proof of Theorem 6, $k = 3$. To improve readability, vertices in $V(G') \setminus V(G)$ and most of the arcs between the leaves of $T_1$ and $T_2$ and the vertices in Red are omitted.

1. We replace every edge $\{v, w\}$ of $G'$ with $v \in \text{Red}$ by an arc $(v, w)$.
2. We add the complete binary tree $T_1$ of lowest height with at least $n$ leaves, with edges directed away from the root. Let $r_1$ denote the root and $l_1^1, \ldots, l_1^n$ leaves of $T_1$.
3. We add the complete binary tree $T_2$ of lowest height with at least $n$ leaves, with edges directed towards the root. Let $r_2$ denote the root and $l_2^1, \ldots, l_2^n$ leaves of $T_2$.
4. For every $1 \leq i \leq n$ we add the arcs $(v_r^i, l_1^i)$ and $(v_r^i, l_2^i)$, running between $G'$ and the trees $T_1$ and $T_2$.
5. We add the arc $(r_2, r_1)$.

This completes the construction of $D$. Next we define the local score function $f$ for $V = V(D)$. Let $\alpha = k - 1$, $\beta = n$ and $\epsilon = 1$. Furthermore, for $v \in V(G')$ we write $N_{G'}(v) = \{ u \in V(G') : \{u, v\} \in E(G') \}$.

(i) For every $v_r^i \in \text{Red}$ we set $f(v_r^i, N_{G'}(v_r^i) \cup l_1^i) = \epsilon$. (ii) For every $v_b^i \in \text{Blue}$ and $\emptyset \neq P \subseteq N_{G'}(v_b^i)$ we set $f(v_b^i, P) = \beta$. (iii) For every $v \in V(D) \setminus (V(G') \cup \{r_2, l_1^1, \ldots, l_1^n\})$ we set $f(v, P_D(v)) = \alpha$. (iv) For the root of $T_2$ we set $f(r_2, P_D(r_2)) = f(r_2, P_D(r_2) \cup \{r_1\}) = \alpha$. (v) For every $l_1^i$ we set $f(l_1^i, P_D(l_1^i)) = f(l_1^i, P_D(l_1^i) \setminus \{v_r^i\}) = \alpha$. (vi) For all the remaining combinations of $v \in V(D)$ and $P \subseteq V(D)$ we set $f(v, P) = 0$.

Evidently $D$ is acyclic and both $D$ and $f$ can be constructed from $G$ in polynomial time. Observe that the super-structure $S_f$ is exactly the skeleton of $D$. Hence, by construction, the degree of every vertex of $S_f$ is bounded by $d + 2$. It remains to show that $G$ has a $k$-Red/Blue non-blocker if and only if $D$ has a $k'$-neighbor $D'$ with a higher score than $D$.

To see this, we first assume that $G$ contains a $k$-Red/Blue non-blocker $S \subseteq \text{Red}$ and $|S| = k$. We obtain $D'$ from $D$ by reversing the $k'$ arcs in $\{ (v_r^i, w) : v_r^i \in S, w \in N_{G'}(v_r^i) \cup \{l_1^i\} \} \cup \{(r_2, r_1)\}$. Note that the reversal of the arc $(r_2, r_1)$ ensures that $D'$ is acyclic, and since $S$ is a $k$-Red/Blue non-blocker in $G$ it follows that the score for every vertex in Blue does not change. Hence $f(D') = f(D) - \alpha + k\epsilon = f(D) + 1 > f(D)$.

To see the reverse direction, note that the vertices in Red are the only vertices of $D$ whose score is not yet maximum. Increasing the score of any of these vertices $v \in \text{Red}$ introduces a cycle that uses only vertices in $V(T_1) \cup V(T_2) \cup \{v\}$. It is easy too see that in order to break this cycle the score for at least one vertex in $V(T_1) \cup V(T_2)$ has to be decreased by $\alpha$ and that all cycles produced in this way can be destroyed by reversing the arc $(r_2, r_1)$. Since $\alpha = (k - 1)\epsilon$ it follows that in order to increase the score for $D$ the score for at least $k$ vertices in Red must be increased to $\epsilon$. Let $S$ be the set of these vertices in Red whose score has been increased in this manner. Since for every vertex in $S$ exactly $k + 1$ arcs need to be reversed and $k' < (d+1)(k+1)$ it follows that $|S| \leq k$ and hence $|S| = k$. Because, $\beta = n\epsilon$ it follows that all vertices in Blue must have kept their score and hence $S$ is a $k$-Red/Blue non-blocker for $G$. $\square$

Theorem 6 provides a surprising contrast to a similar study of $k$-local search for MAX-SAT where the problem is fixed-parameter tractable for instances of bounded degree (Szeider 2009). A possible explanation for the surprising hardness of $k$-$\mathbb{O}$-LOCAL SEARCH BAYESIAN STRUCTURE LEARNING could be that, in contrast to MAX-SAT, a global property of the entire instance (acyclicity) must be checked.

| network | $n$ | $m$ | $w$ | $d$ |
|---|---|---|---|---|
| link | 724 | 1125 | 16 | 17 |
| alarm | 37 | 46 | 3 | 6 |
| carpo | 61 | 74 | 4 | 12 |
| barley | 48 | 84 | 6 | 8 |
| hailfinder | 56 | 66 | 3 | 17 |
| diabetes | 413 | 602 | 4 | 24 |
| insurance | 27 | 52 | 6 | 9 |
| win95pts | 76 | 112 | 5 | 10 |
| mildew | 35 | 46 | 3 | 5 |
| munin1 | 189 | 282 | 11 | 15 |
| munin2 | 1003 | 1244 | 6 | 30 |
| munin3 | 1044 | 1315 | 8 | 69 |
| munin4 | 1041 | 1397 | 8 | 69 |
| pigs | 441 | 592 | 9 | 41 |
| water | 32 | 66 | 7 | 8 |

Table 1: Bayesian networks from http://compbio.cs.huji.ac.il/Repository/. $n$ = number of nodes, $m$ = number of edges, $w$ = upper bound on the treewidth, $d$ = maximum degree. All parameters refer to the skeleton of the network.

## 7 Conclusion

We have studied the computational complexity of exact Bayesian Structure Learning under graph-theoretic restrictions on the super-structure. Our results show that exact learning is linear-time tractable if the super-structure has bounded treewidth and bounded maximum degree, but none of the two restrictions can be dropped without loosing linear time tractability (or uniform polynomial-time tractability). Our algorithm is based on dynamic programming along a tree decomposition of the super-structure. We have focused on theoretical worst-case complexity results, leaving an empirical evaluation of the algorithm on real-world data for future research. As a first step in that direction we have computed treewidth and maximum degree of the skeletons of some well-known benchmark networks and found relatively small numbers, see Table 1. We take this as an encouraging indication that from a practical point of view it makes sense to consider super-structures of small treewidth and small maximum degree. In fact, it is desirable to learn networks of small treewidth and small maximum degree as such networks allow efficient reasoning. On the theoretical side we offer as an objective for future research the identification of other graph-theoretic parameters that allow efficient exact structure learning. In particular, it would be interesting to identify parameters that, in contrast to treewidth and maximum degree, separate the (parameterized) complexities of finding globally optimal networks from improving networks locally by $k$-neighborhood local search.